%% file: main.tex
\documentclass[10pt,twocolumn,letterpaper]{article}

\usepackage{cvpr}              % To produce the CAMERA-READY version

\definecolor{cvprblue}{rgb}{0.21,0.49,0.74}
\definecolor{dtdark}{gray}{0.5}
\definecolor{dt}{gray}{0.6}
\usepackage[pagebackref,breaklinks,colorlinks,allcolors=cvprblue]{hyperref}
\usepackage{multirow}
\usepackage{tikz}
\usepackage{utfsym}
\usepackage{comment}
\usepackage{color}

\def\paperID{***} % *** Enter the Paper ID here
\def\confName{CVPR}
\def\confYear{2025}

\newcommand{\dataset}{\textit{Insight-UI Dataset}\xspace}
\newcommand{\model}{\textit{Falcon-UI}\xspace}
\title{Falcon-UI: Understanding GUI Before Following User Instructions}

\author{\textbf{Huawen Shen\textsuperscript{1,5}\ \ Chang Liu\textsuperscript{3}\ \ Gengluo Li\textsuperscript{1,5}\ \ Xinlong Wang\textsuperscript{4}\ \ Yu Zhou\textsuperscript{2}\ \ Can Ma\textsuperscript{1}\ \ 
Xiangyang Ji\textsuperscript{3}} 
\\
\textsuperscript{1} Institute of Information Engineering, Chinese Academy of Sciences\\
\textsuperscript{2} VCIP \& TMCC \& DISSec, College of Computer Science, Nankai University\\
\textsuperscript{3} Department of Automation, Tsinghua University\\
\textsuperscript{4} Beijing Academy of Artificial Intelligence\\
\textsuperscript{5} School of Cyber Security, University of Chinese Academy of Sciences\\
\tt{\{shenhuawen, ligenglo, macan\}@iie.ac.cn, yzhou@nankai.edu.cn}\\
\tt{\{liuchang2022, xyji\}@tsinghua.edu.cn, wangxinlong@baai.ac.cn}
}

\begin{document}

\twocolumn[{
\maketitle\centering
\captionsetup{type=figure}
\includegraphics[width=\linewidth]{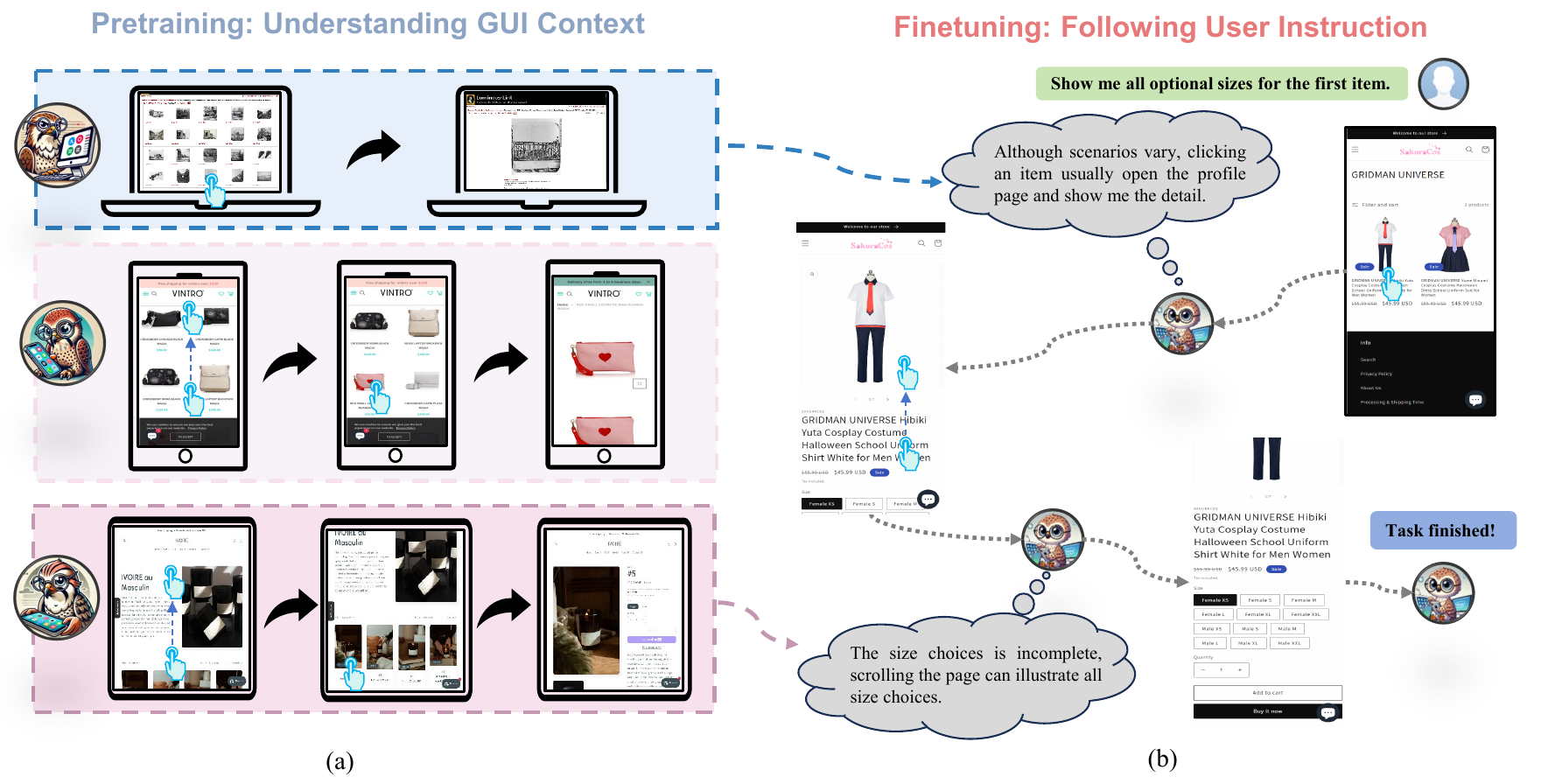}\vspace{-2mm}
\captionof{figure}{(a) Different GUI scenarios share similar operation logic, pretraining the GUI agent on diverse GUI examples can equip models with a broader understanding of GUI contexts. (b) A model with enhanced GUI context knowledge can leverage similarities across scenarios, reducing optimization complexity in downstream tasks.}
\label{fig:paradigm}
}]

\maketitle

\begin{tikzpicture}[remember picture,overlay,shift={(current page.north west)}]
\node[anchor=north west, xshift=2.5cm, yshift=-3.4cm]{\scalebox{-1}[1]{\includegraphics[width=0.8cm]{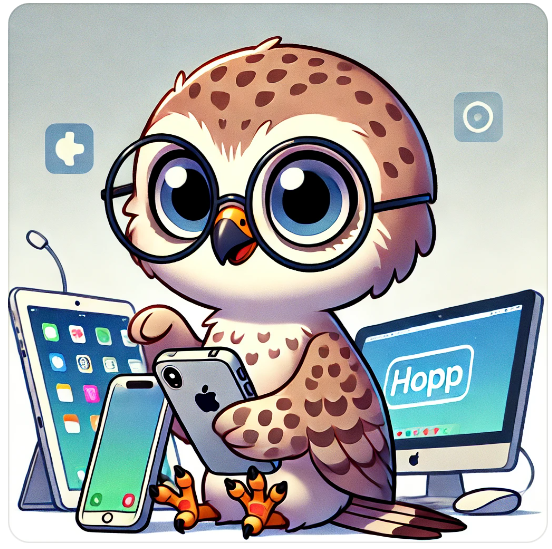}}};
\end{tikzpicture}

\input{sec/0_abstract}

\input{sec/1_introduction}
\input{sec/2_related}
\input{sec/3_method}
\input{sec/4_experiment}
\input{sec/5_conclusion}
% \newpage
{
    \small
    \bibliographystyle{ieeenat_fullname}
    \bibliography{main}
}

\input{sec/X_supplementary}

\end{document}

%% file: sec/0_abstract.tex
\begin{abstract}
Pursuing human-like interaction for Graphical User Interface (GUI) agents requires understanding the GUI context and following user instructions.
However, existing works typically couple these two aspects and focus more on instruct-following abilities, while ignoring the importance of understanding the GUI context.
In this paper, we introduce an instruction-free GUI navigation dataset, termed \dataset, to enhance model comprehension of GUI environments. 
\dataset is automatically generated from the Common Crawl corpus, simulating various platforms -- including iOS, Android, Windows, and Linux -- across multiple resolutions on 312K domains.
Although GUI interactions vary by context, diverse interfaces share common internal patterns, such as clicking an item to view its details. 
It implies the feasibility of independent GUI operation learning, followed by joint optimization with instruction tuning.
Thereby, we develop the GUI agent model \model \footnote{The falcon is instinctively driven to hunt, and the falconer harnesses this instinct. Falcon-UI is pretrained to develop GUI comprehension and is subsequently activated by downstream instruction.}, which is initially pretrained on \dataset and subsequently fine-tuned on Android and Web GUI datasets, including AITW, AITZ, Android Control, and Mind2Web.
With 7 billion parameters, \model achieves accuracy comparable to the 72 billion-parameter Qwen2VL on AITZ,
validating the alignment between GUI context comprehension and agent performance.
Our code and dataset will be open-sourced.
\end{abstract}

%% file: sec/1_introduction.tex
\section{Introduction}
\label{sec:introduction}

Modern operating systems primarily utilize Graphical User Interfaces (GUIs) for display, and autonomous interaction with GUIs is proposed to decrease human workload~\cite{Appagent,AutoGPT}.
Early GUI agents~\cite{Wob,TextAgent} typically process structured inputs from GUI source code, such as HTML, and invoke predefined system functions, which limits their generalization across non-adapted systems and applications.
The emergence of large vision-language models (LVLMs)~\cite{MiniGPT,mplug-owl} has prompted the development of GUI agents that interact with GUIs using visual inputs and coordinated outputs~\cite{SeeClick,CogAgent}, thereby mimicking genuine human interactions.
By eliminating the reliance on system API, GUI agents can be applied to a wide range of scenarios.

To function as a GUI Agent, the model requires two essential capabilities: (1) recognizing and interpreting the GUI context, and (2) comprehending user instructions to predict appropriate actions for GUI interaction.
In contrast to typical visual recognition tasks with static visual input, GUI environments are dynamic~\cite{Visualwebarena,AndroidWorld}, adapting based on model predictions.
For the model to predict actions effectively, it must not only process the visual layout but also identify the role and potential responses associated with each element.

However, previous approaches mainly focus on constructing a dataset with specific user instructions:
(1) Engage multiple human annotators to carry out specific tasks based on predefined instructions~\cite{UIBert,motif}. This process is time-consuming and costly, limiting the quantity and diversity of instructions and GUI scenarios, and restricting the generalization to unseen scenarios.
(2) Leverage an LVLM to refine instructions for existing data~\cite{AITZ,Ferret-UI_2}. This approach introduces complicated procedures, and the data quality is highly relevant to the LVLM's proficiency in GUI navigation.
(3) Employ heuristic rules to extract auxiliary information from the original GUI source code, such as the \textit{alt} tag in HTML~\cite{Spotlight}. However, not all scenarios provide auxiliary information, and some developers either leave it as default or include irrelevant fields, leading to unexpected errors and limited diversity within these datasets.

Given the challenge of creating diverse, high-quality user instructions, we propose decoupling instructions from the GUI context.
As shown in \cref{fig:paradigm}(a), although from different platforms and fields, these episodes share similar interaction logic -- clicking an element will reveal its details.
Once equipped with such GUI knowledge, the agent can smoothly adapt to downstream scenarios where a user instruction is given, as shown in \cref{fig:paradigm}(b).

As the initial step in this instruction-free training paradigm, we develop an instruction-free dataset, \dataset.
To ensure stable performance across diverse scenarios, \dataset is designed to cover various fields and complex appearances.
Specifically, we download raw data from Common Crawl, render pages, and simulate user actions using the Browser API. We capture page screenshots, visible node information, and interactions during the simulation.
\dataset ultimately contains 434K episodes from 312K domains, with 1,456K images in total.\textbf{
\dataset is entirely auto-generated, requiring no human annotators or LVLMs.}
It can easily expand to any GUI scenario, with only screen capture and user action simulation privileges required -- exactly those needed for GUI agent inference on devices.
Thus, for any scenario where the GUI agent can be employed, a corresponding dataset can be generated, indicating the potential to further optimize on a specific domain.

Finally, we introduce a GUI agent model, \model, to validate the effectiveness of our proposed instruction-free pretraining. \model is designed to take screenshots as input and perform basic GUI actions like clicking and typing.
\model is first enhanced with GUI understanding capability using \dataset, and then fine-tuned on downstream mobile and web GUI agent tasks, including AITW~\cite{AITW}, AITZ~\cite{AITZ}, Android Control~\cite{AndroidControl}, and Mind2Web~\cite{Mind2Web,SeeClick}.
Notably, with only 7 billion parameters, \model achieves accuracy comparable to the 72 billion-parameter Qwen2VL on AITZ.

Our contributions can be summarized as follows:

\begin{itemize}
    \item We highlight the challenge of creating diverse, high-quality GUI instruction datasets and propose a new training paradigm that decouples the user instruction from GUI pretraining.
    \item We develop \dataset as the initial instruction-free dataset, which is fully auto-generated and adaptable to various scenarios.
    \item We propose a general GUI agent, \model, which, after being enhanced by \dataset, demonstrates impressive performance on multiple datasets, highlighting the importance of equipping models with GUI understanding capabilities.
\end{itemize}

%% file: sec/2_related.tex
\section{Related Work}
\label{sec:related}

\begin{figure*}[t]
  \centering
   \includegraphics[width=\linewidth]{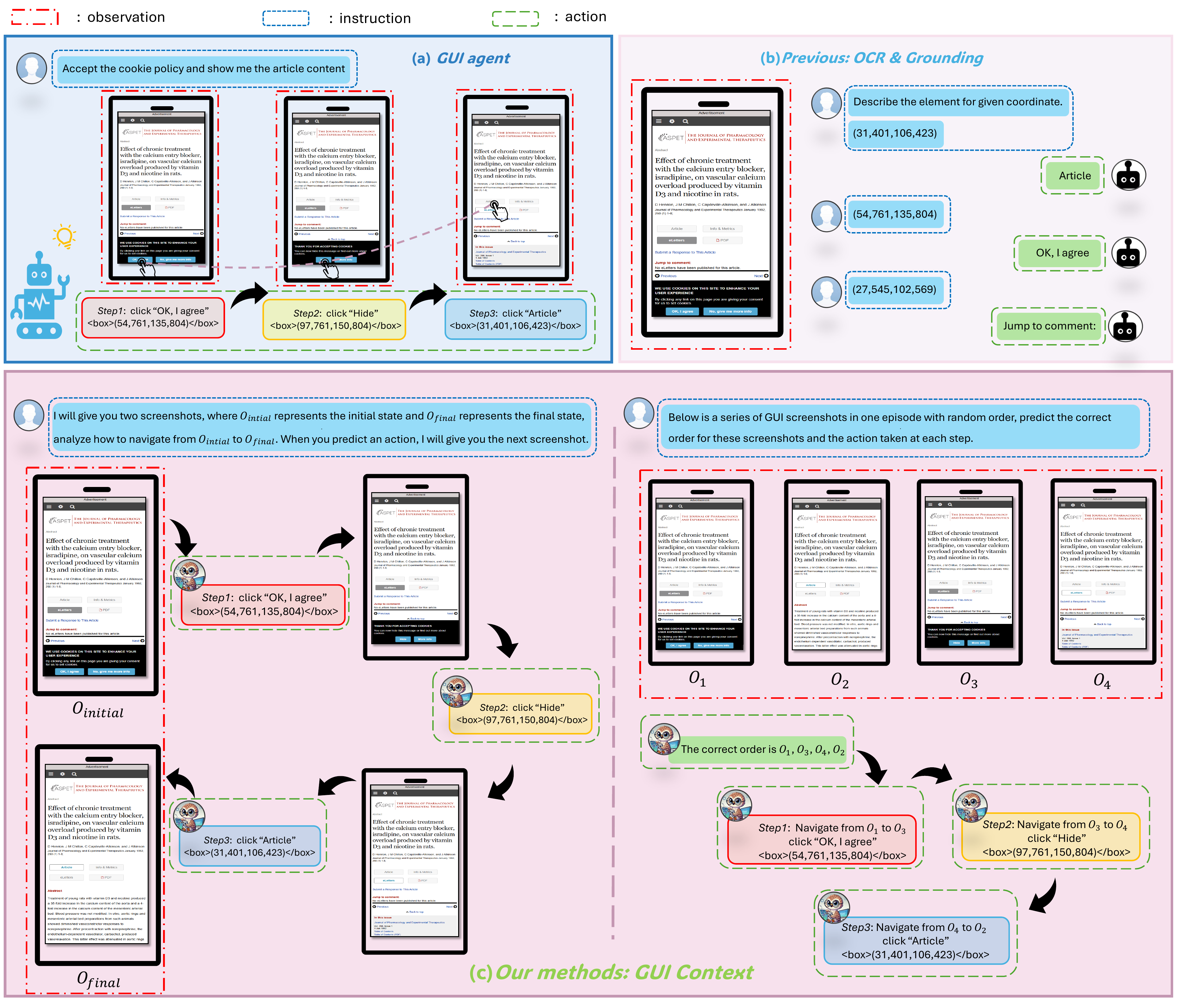}
   \caption{Comparison of different task paradigms. (a) The dynamic environment characterizes the GUI agent task, and the observation will change according to the action prediction. (b) Previous GUI pretraining datasets mainly focus on single-image element grounding, ignoring the relationship among different GUI observations. (c) Our proposed paradigm takes the observation changes into consideration.}
   \label{fig:comparison}
\end{figure*}

\subsection{Large Vision-Language Models}

Inspired by the success of large language models (LLMs), researchers are exploring the integration of vision modality into LLMs. Flamingo~\cite{Flamingo} first proposes to integrate vision embedding into the representations of language model with cross-attention layers. LLaVA~\cite{LLaVA} uses a simple MLP structure to concatenate vision tokens with text sequence. InstructBLIP~\cite{InstructBLIP} applies Q-Former to better compress vision information. When facing complex scenarios where visual details are required, LLaVA-NeXT~\cite{LLaVA-Next} applies the ``AnyRes" technique to fit dynamic high resolution. Recent works, like DeepSeek-VL~\cite{DeepSeek-VL} and Qwen2VL~\cite{Qwen2VL}, curate large quantities of training data and show a comparative performance with commercial models like GPT-4o.

\subsection{GUI Agent}

Following the progress of large Vision-Language Models (LVLM), GUI agent is proposed to simulate human operation with image-only input.
Early attempts~\cite{Synapse,NLPAgent1} try to prompt LLM to perform navigation tasks.
Auto-UI combines BLIP2~\cite{BLIP2} vision encoder with FLAN-Alpaca~\cite{FLAN-Alpaca} to build a multi-modal solution, bypassing the need for environment parsing or reliance on application-dependent APIs.
CoCoAgent~\cite{CoCoAgent} integrates detailed elements layout to further improve GUI perception capability, and decomposes the action prediction into type prediction and target selection.
To better utilize vision modality and adapt the agent to broader scenarios, LVLM is applied to achieve the GUI navigation.
AppAgent series~\cite{Appagent,AppAgentv2} apply GPT4 to explore the Android APP to create a reference document and utilize this document to navigate the APP.
CogAgent~\cite{CogAgent} enhances the CogVLM~\cite{CogVLM} with larger resolution input to augment GUI navigation, supporting both mobile devices and desktop webpages.
SeeClick~\cite{SeeClick} identifies the GUI grounding as the core capability for developing a GUI agent, and the model is enhanced on GUI grounding data to locate screen elements based on user instruction.
MobileVLM~\cite{MobileVLM} select 49 Chinese APPs to formulate the navigation graph within specific APP.
Ferret-UI~\cite{FerretUI} applies ``any resolution" to magnify the GUI detail and builds a GUI dataset from existing Android~\cite{Rico} and IOS~\cite{AMP} datasets. It collects fine-grained element annotation using a pretrained pixel-based UI detection model and constructs detailed descriptions with GPT-4.
Ferret-UI 2~\cite{Ferret-UI_2} further extends to multiple platforms.

%% file: sec/3_method.tex
\section{Method}
\label{sec:method}

\subsection{GUI Context Understanding}

As shown in \cref{fig:comparison}(a), a typical GUI agent task can be framed as a multi-modal, multi-turn interaction consisting of three key components: instruction, observation, and action.
The instruction $I$ defines the ultimate goal of the task, specifying what the user intends the agent to achieve.
At each time step $t$, the agent receives the observation $o_t$, represented as a screenshot from the GUI environments.
The model then predicts the action $a_t$ based on the current context.
The GUI agent task involves training a model that maximizes the probability of $\pi (a_t | I, [h_1, h_2, ..., h_{t-1}], o_t)$, where $h_t = (o_t, a_t)$.

However, previous GUI agent pretraining datasets typically use single-image formats, as illustrated in \cref{fig:comparison}(b).
In these datasets, the observations are static.
Due to the challenge of constructing user instructions, two tasks are commonly included in these datasets: a low-level elementary task and a high-level reasoning task.
The low-level elementary task often involves element grounding, where the model is asked to locate OCR text or icon types. Annotations typically come from source code metadata~\cite{SeeClick}.
The high-level reasoning task typically involves either image captioning or Visual Question Answering (VQA), where human annotators~\cite{Rico} or existing LVLMs~\cite{FerretUI} are prompted to describe the functionality of specific elements or entire pages.
Although some datasets involve multi-turn conversations, the images remain unchanged, and the instructions within each image are independent. Deleting or reordering the instructions does not affect the task.
Finally, the model is optimized as $\pi (a | I, o)$.
A model trained in this manner lacks the knowledge of dynamic GUI environments, making it difficult to understand the next state when interacting with specific elements.

Here, we propose to apply the GUI context as the pretraining scenario. The GUI context consists of interleaved observations and actions, which represent the transition logic between different GUI pages.
As shown in \cref{fig:comparison}(c), given a sample of the GUI context, the model is tasked with predicting how to navigate from one observation to another.
The model is optimized to $\pi (a_t | o_{target}, [h_1, h_2, ..., h_{t-1}], o_t)$. 
To achieve this, the model must understand the relationships between different observations and the functionality of each element.
This paradigm offers three key advantages:

\begin{itemize}
    \item \textbf{GUI context understanding.} The observations in this paradigm are closely related. The model learns the consequences of interacting with the specific element and how to navigate to the expected states.
    \item \textbf{High data quality.} In the data collection procedure, we only need to record the observation after interacting with an element, eliminating the need to extract potentially inaccurate metadata from source code or external tools.
    \item \textbf{Cost-effective construction procedure.} By avoiding human annotators and external tools like LVLMs, we only run a light simulator to collect the GUI context data.
\end{itemize}

We use two pretraining formats, as illustrated in \cref{fig:comparison}(c). 
The first format begins with the initial and final observations. Each time the model predicts an action, a new observation is provided, simulating real-world GUI agent tasks.
The second format presents all observations initially, and the model is tasked with analyzing the order of these observations and the actions taken at each step.
As described in \cref{data_construct}, we use an automatic random walk procedure to construct our dataset, which may result in irrelevant actions in long GUI contexts, making it impossible for the model to predict it.
To mitigate this, we first train an agent using the first format with 2000 pretraining examples. The trained model is then used to perform inference on the whole pretraining dataset. The half examples with the largest loss are selected for the second format.
Additional examples and details are provided in the Appendix.

\begin{figure*}[t]
  \centering
   \includegraphics[width=\linewidth]{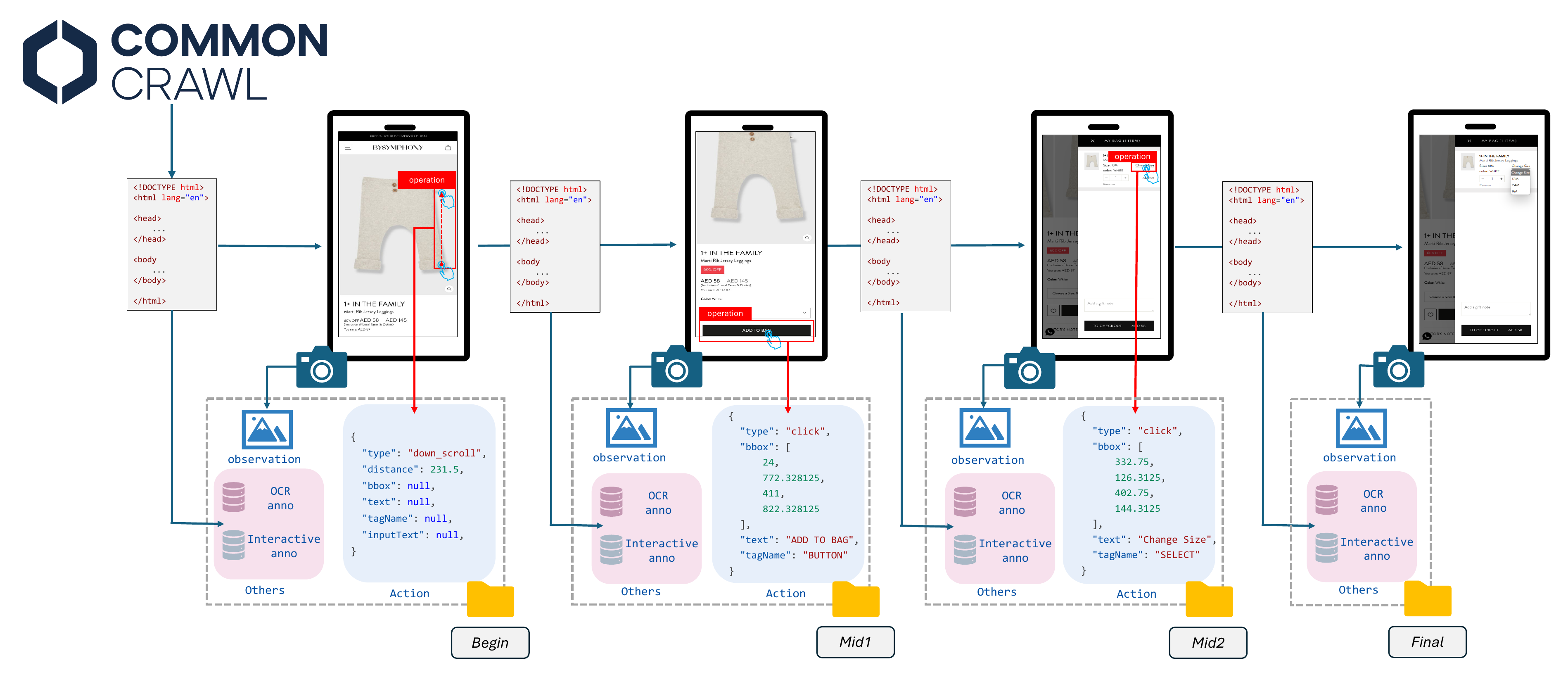}
   \caption{Data collection pipeline for \dataset. After rendering a page from HTML, an action is selected to simulate human operation. We record all observation and action information during the procedure.}
   \label{fig:data_collect}
\end{figure*}

\begin{table*}[]
\centering
\begin{tabular}{lccccccl}
\toprule
\multirow{2}{*}{Dataset}   & \multirow{2}{*}{Episodes}  & \multirow{2}{*}{Images}  & Domains/   & Multi-image  & Annotator  & Metadata   & \multirow{2}{*}{Platform}   \\ 
          &          &         & Apps   & Context  & free & free        \\ \midrule
Screen2Words~\cite{Screen2words}   & -          & 22 K          & 6.3 K                  & \usym{2717}         & \usym{2717}         & \usym{2713}         & Mobile               \\
Rico~\cite{Rico}                   & -          & 72 K          & 9.7 K                  & \usym{2717}         & \usym{2717}         & \usym{2713}         & Mobile               \\
AMP~\cite{AMP}                     & -          & 81 K          & 4.2 K                  &  \usym{2717}        & \usym{2717}         & \usym{2717}         & Mobile               \\
SeeClick~\cite{SeeClick}           & -          & 300 K         & -                      &  \usym{2717}        & \usym{2713}         & \usym{2717}         & Desktop              \\
Ferret-UI~\cite{FerretUI}          & -          & 111 K         & -                      &  \usym{2717}        & \usym{2713}         & \usym{2717}         & Mobile               \\ \midrule
\dataset                         & 434 K       & 1,456 K       & 312 K                  & \usym{2713}          & \usym{2713}         & \usym{2713}         & Mobile \& Desktop    \\ \bottomrule
\end{tabular}
\caption{Dataset statistics of \dataset and previous GUI pretraining datasets. \dataset is sourced from diverse domains and offers different resolution screenshots, enhancing the generalization on different downstream scenarios.}
\label{tab:dataset}
\end{table*}

\begin{table*}[]
\centering
\begin{tabular}{cccccccc}
\toprule
Model                                      & Vision-Only                  & General        & Install        & GoogleApps     & Single                                  & WebShop        & Overall        \\ \midrule
BC-history \cite{AITW}                      &  \usym{2717}                 & 63.70           & 77.50           & 75.70           & 80.30                                    & 68.50           & 73.10           \\
LLaMa 2-7B \cite{LLaMa2}$^\dag$             &  \usym{2717}                 & 28.56           & 35.18          & 30.99          & 27.35                                   & 19.92          & 28.40          \\
MobileAgent \cite{MobileAgent-Finetune}     &  \usym{2717}                 & 55.80           & 74.98          & 63.95          & 76.27                                   & 63.61          & 66.92          \\
CoCo-Agent \cite{CoCoAgent}                 &  \usym{2717}                 & 69.92          & 80.60          & 75.76          & 88.81                                   & 74.02          & 77.82          \\ \midrule
GPT-4o$^\ddag$                             &  \usym{2713}                 & 47.06          & 49.12          & 52.30          & 80.28                                   & 46.42          & 55.02          \\
OmniParser \cite{OmniParser}                &  \usym{2713}                 & 48.30           & 57.80           & 51.60           & 77.40                                    & 52.90           & 57.70           \\
Auto-UI \cite{AutoUI}                       &  \usym{2713}                 & 68.24          & 76.89          & 71.37          & 84.58                                   & 70.26          & 74.27          \\
SeeClick \cite{SeeClick}                    &  \usym{2713}                 & 67.60           & 79.60           & 75.90           & 84.60                                    & 73.10           & 76.20           \\
MobileVLM \cite{MobileVLM}                  &  \usym{2713}                 & 69.58          & 79.87          & 74.72          & 81.24                                   & 71.70          & 74.94          \\
CogAgent \cite{CogAgent}                    &  \usym{2713}                 & 65.38          & 78.86          & 74.95          & \textbf{93.49}                          & 71.73          & 76.88          \\ \midrule
\textbf{Ours} &  \usym{2713}            & \textbf{70.48} & \textbf{81.63} & \textbf{76.08} & \textit{\underline{86.69}} & \textbf{74.23} & \textbf{77.82} \\ \bottomrule
\end{tabular}
\caption{\textbf{Comparison on Android in the Wild (AITW).} For the purely vision-based input model, the best results are indicated in \textbf{bold}, and the second-best results are marked in \textit{\underline{underline}}. $\dag$ denotes results from \cite{AutoUI}, while $\ddag$ denotes results from \cite{MobileVLM}.}
\label{tab:AITW}
\end{table*}

\subsection{Dataset Construction}
\label{data_construct}

Our data collection procedure is illustrated in \cref{fig:data_collect}. 
% we apply Common Crawl as our beginning corpus. 
After downloading the raw data from Common Crawl, we filter out duplicates, unreachable URLs, and non-English, NSFW websites based on the main content of each page. 
We apply the puppeteer library\footnote{https://github.com/puppeteer/puppeteer} to simulate a browser environment. To ensure the diversity of our curved dataset, we simulate different environments by puppeteer API with different resolutions and platform types, including mobile, tablet, and desktop devices on different operating systems such as Android, iOS, Windows, and Linux.
For each GUI page, we record the screenshot and corresponding node information. Different from text-modality datasets that download the whole page, we only record the visible elements, simulating real interaction like human beings.

Then a random interaction is performed on the page, and a new observation will be obtained.
The record-interact loop is repeated until a predefined condition is met, such as five interactions are performed or a time-out error encounters. The types of interactive actions include:

\begin{itemize}
    \item \textbf{Click} operation is the most frequently performed operation in a GUI agent. Only the visible and clickable elements will be selected.
    \item \textbf{Scroll} operation is used to look through the invisible area, and we record the scroll direction and distance.
    \item \textbf{Input} operation is commonly seen in the search bar and forms. Similar to the click operation, only the visible and inputtable elements will be selected.
\end{itemize}

Some heuristic rules are applied to ensure better action choosing, and more details can be found in our Appendix.
After collection, we clean the data, removing the episodes with blank pages or duplicate pages.

We rent the computing virtual machine on Google Cloud to execute our simulator, render the GUI, and collect all needed data.
Finally, 1,019K episodes and 3,431K images are collected, after post-processing, 434K episodes and 1,456K images remain.
A simple comparison with previous GUI pretraining datasets is illustrated in \cref{tab:dataset}.
We record all visible interactive elements and character-level OCR annotation with a hierarchical document for each screenshot. We hope these raw data could benefit the community as better post-processing could be applied in the future.
\textbf{The whole data collection procedure costs \$200.}
For a comparison, Rico\cite{Rico} spends \$19,200 for crowdsource annotations.

\subsection{Model Optimization}
\label{model_optimization}

To maintain the model's instruction-following abilities, we incorporate general vision-language data into the pretraining stage.
Specifically, our training data consists of three components:

\begin{itemize}
    \item Our proposed \dataset is used to embed GUI knowledge into the model.
    \item M4-Instruct~\cite{M4-Instruct} enhances multi-image processing abilities. We remove the samples which contain more than 5 images or the image resolution is smaller than 100.
    \item 158k GPT-generated instruction data proposed by LLaVA~\cite{LLaVA} is included to retain complex instruction-following capabilities.
\end{itemize}

We finally mix these datasets to create a 1M-sample dataset to pretrain \model.
\model is continually pre-trained on Qwen2-VL-7B, which demonstrates extraordinary vision-language reasoning capabilities and accepts any resolution vision input.
We train Qwen2VL on the mixed dataset for 1 epoch to enhance the GUI understanding ability. We froze its vision encoder throughout the training process, as in our preliminary experiments, joint training with the ViT did not yield improvements. More details can be found in our Appendix.

%% file: sec/4_experiment.tex
\section{Experiment}
\label{sec:experiment}

To evaluate the effectiveness of GUI context pretraining, we conduct extensive experiments on a broad range of datasets. First, we conduct evaluations on GUI agent benchmarks for both mobile and desktop scenarios. Then further analysis is conducted to investigate the effect of different pretraining data.

\subsection{Main Result}

\noindent\textbf{{Evaluation on AITW.}} Android In The Wild (AITW) \cite{AITW} is the pioneer of GUI Agent, which consists of 715K episodes from Android devices, with template-based instruction and action sequences from human annotators. 
AITW includes eight types of devices, ranging from Pixel 2 XL to Pixel 6, each with different screen resolutions.
AITW divides the data into five subsets: General, Install, Google Apps, Single, and WebShopping, and we jointly fine-tune \model on all the subsets.
Due to the severe imbalance of data distribution, for the GoogleApps subset, we randomly sample 10\% training data following previous works \cite{CogAgent}.
The results on shown in \cref{tab:AITW}. \model achieves state-of-the-art performance compared to previous vision-based models, gaining an overall improvement of 1\% over CogAgent.
As for the model with auxiliary information, \model also achieves comparable accuracy with CoCo-Agent.

\begin{table}[]
\centering
\begin{tabular}{cccc}
\toprule
\#                          & Model       & Exact          & Type           \\ \midrule
\multirow{2}{*}{Few-shot}   & CogAgent \cite{CogAgent}$^\dag$    & 53.3           & 72.6          \\
                            & GPT-4o$^\ddag$      & 35.3           & 70.0           \\ \midrule
\multirow{3}{*}{Fine-tuned} & Auto-UI \cite{AutoUI}$^\dag$     & 47.7           & 83.0          \\
                            & Qwen2-VL-72B \cite{Qwen2VL} & \textbf{72.1}  & \textbf{89.6}  \\ 
                            & \textbf{Ours}        & \textit{\underline{69.1}}  & \textit{\underline{84.7}} \\ \bottomrule
\end{tabular}
\caption{\textbf{Comparison on Andoird in the Zoo (AITZ)}. Our 7 billion-parameter model reaches comparable accuracy with 72 billion-parameter Qwen2-VL. $\dag$ denotes results from \cite{AITZ}, while $\ddag$ denotes results from \cite{Qwen2VL}.}
\vspace{-4pt}
\label{tab:AITZ}
\end{table}

\noindent\textbf{{Evaluation on AITZ.}} Android In The Zoo (AITZ) \cite{AITZ} is a fine-grained version of AITW. 
It locates the error annotation in AITW where the observed screenshots do not match the instructions, or the annotated sequences fail to complete the instruction. 
AITZ samples out 7,180 episodes from AITW, deletes the wrong annotations, and re-annotates the data to offer detailed and diverse instruction with human annotators and LLM. Finally, AITZ constructs a clean dataset composed of 1,998 training samples and 506 test samples.
As shown in \cref{tab:AITZ}, \model achieves state-of-the-art performance over Auto-UI and CogAgent. \model is continually pre-trained on Qwen2-VL-7B, with the same architecture design, it achieves comparable accuracy with the 72 billion-parameter model, validating the effectiveness of GUI context understanding pretraining.

\begin{table}[]
\centering
\begin{tabular}{cccc}
\toprule
\#                          & Model          & high-level & low-level \\ \midrule
\multirow{2}{*}{Zero-shot}  & Gemini 1.5 Pro & 24.4       & 50.2      \\
                            & GPT-4         & 32.1       & 51.7      \\ \midrule
Few-shot                    & Gemini 1.5 Pro & 39.5       & 53.3      \\ \midrule
\multirow{2}{*}{Fine-tuned} & PaLM 2S~\cite{Palm_2}        & \underline{\textit{70.8}}       & \underline{\textit{83.2}}      \\
                            & \textbf{Ours}           & \textbf{72.7}       & \textbf{86.6}       \\ \bottomrule
\end{tabular}
\caption{\textbf{Comparison on Android Control}. Other models' results come from \cite{AndroidControl}.}
\label{tab:AndroidControl}
\end{table}

\noindent\textbf{{Evaluation on Android Control.}} Given the limited number of apps and instructions in previous benchmarks, Android Control \cite{AndroidControl} expands to cover a broader range of scenarios.
The initial task descriptions for Android Control are generated using LLMs and encompass 40 distinct categories.
Human annotators may use any app of their choice to complete the tasks and are required to provide high-level natural language descriptions for the whole episode.
Before executing each action, annotators must provide a brief natural language description of the action they plan to take, thereby generating low-level instructions for each step.
Data collection for Android Control spans over a year. Finally, Android Control contains 15,283 unique tasks across 833 Android apps, making it one of the highest-quality Android GUI agent benchmarks available today.
The results are illustrated in \cref{tab:AndroidControl}, \model achieves state-of-the-art performance on both high-level and low-level instruction, validating the pretraining for GUI context understanding could benefit diverse downstream scenarios.

\begin{table}[]
\centering
\begin{tabular}{ccccc}
\toprule
\multirow{2}{*}{Model} & Cross- & Cross-   & Cross-  & \multirow{2}{*}{overall} \\
                       & Task  & Website & Domain &     \\ \midrule
\multicolumn{5}{l}{\color{dtdark}\it{\textbf{input with element candidates}}} \\ \midrule
\color{dtdark}GPT-3.5  & \color{dtdark}17.4 & \color{dtdark}16.2    & \color{dtdark}18.6   &  \color{dtdark}17.4       \\
\color{dtdark}GPT-4     & \color{dtdark}36.2 & \color{dtdark}30.1    & \color{dtdark}26.4   & \color{dtdark}30.9        \\
\color{dtdark}CogAgent~\cite{CogAgent} & \color{dtdark}62.3 & \color{dtdark}54.0    & \color{dtdark}59.4   & \color{dtdark}58.2        \\ \midrule
\multicolumn{5}{l}{\it{\textbf{input with pure-vision modality}}} \\ \midrule
Qwen-VL~\cite{QwenVL}  & 13.3 & 9.2     & 12.0   & 11.5        \\
SeeClick~\cite{SeeClick} & 25.5 & 16.4    & 20.8   & 20.9        \\
\textbf{Ours}     & \textbf{31.7} & \textbf{25.8}    & \textbf{25.2}   &  \textbf{27.6}        \\ \bottomrule
\end{tabular}
\caption{\textbf{Comparison on Mind2Web.} We adopt the screenshots from SeeClick \cite{SeeClick}.}
\label{tab:Mind2Web}
\end{table}

\noindent\textbf{{Evaluation on Mind2Web (Multi-modal).}} 
Mind2Web \cite{Mind2Web} is a web-based dataset collected from desktop devices, featuring a significantly higher resolution compared to mobile datasets.
Initially, Mind2Web was designed for pure language models, with HTML as the sole input format.
Later, a multi-modal version\footnote{https://huggingface.co/datasets/osunlp/Multimodal-Mind2Web} was introduced, which captures the entire webpage as screenshots, resulting in a resolution that exceeds 10,000.
Previous methods, whether language-based (e.g., LLaMA) or vision-based (e.g., CogAgent), typically first extract candidates from simplified HTML, then apply the agent to select the target element from candidates.
This complicates the evaluation of whether the agent understands the visual GUI context or relies solely on its language capabilities.
Therefore, SeeClick \cite{SeeClick} constructs a multi-modal version of Mind2Web using the raw data, capturing only the visible portion as a screenshot, and directly predicting the interaction coordinates, simulating real user interaction.
We adopt the SeeClick settings -- all training and testing data for Mind2Web are sourced from the SeeClick official code repository\footnote{https://github.com/njucckevin/SeeClick}. Additionally, we evaluate our performance using the SeeClick methodology: the model predicts actions based on coordinates, and a prediction is deemed correct if the predicted coordinate falls within the target element's bounding box.
As shown in \cref{tab:Mind2Web}, \model significantly outperforms the baseline SeeClick, demonstrating its ability to generalize to desktop devices.

\begin{figure}[t]
  \centering
   \includegraphics[width=\linewidth]{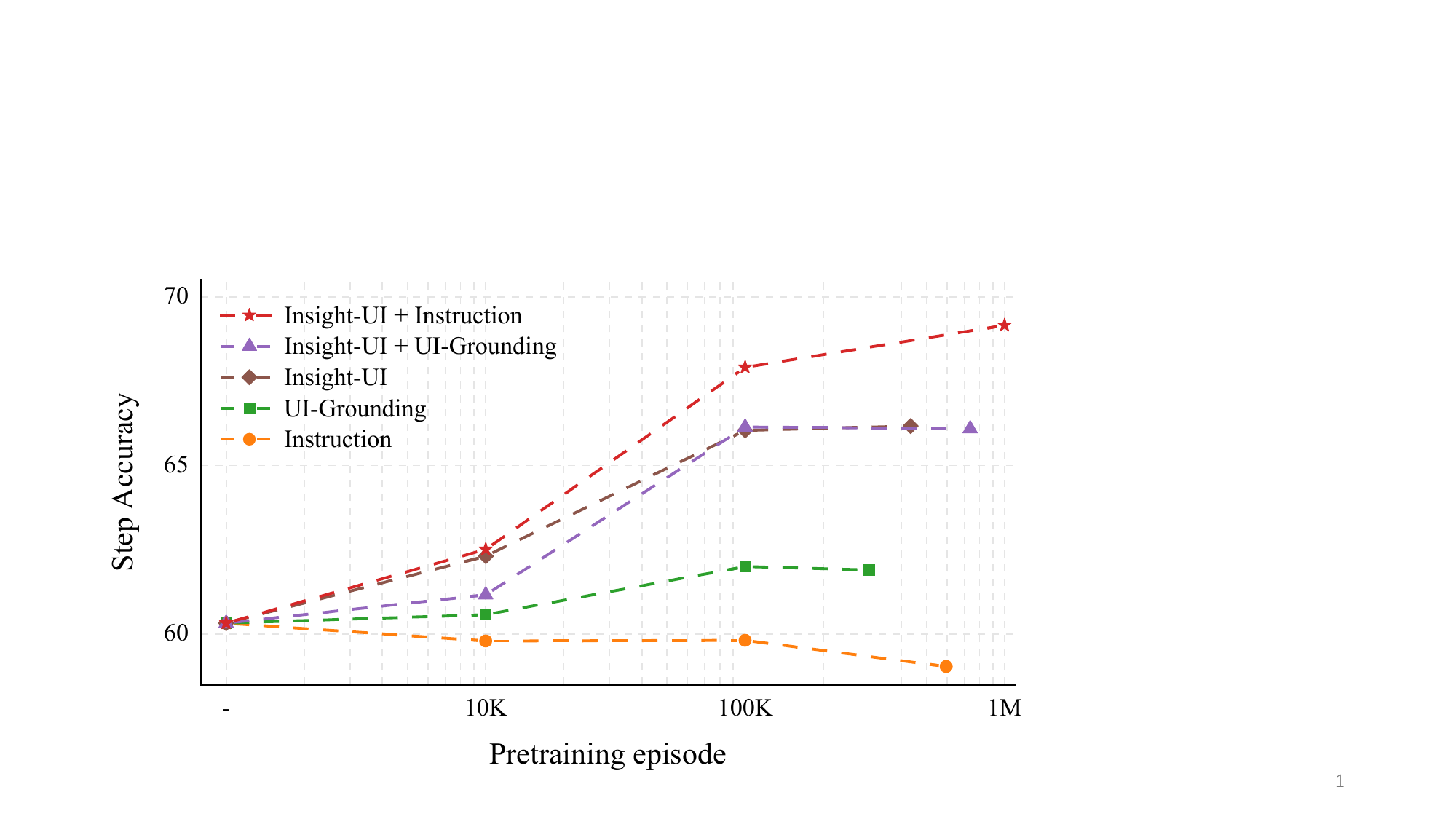}
   \caption{After pretraining on different data combinations, the model is fine-tuned and evaluated on AITZ.}
   \label{fig:data_mixture}
\end{figure}

\begin{figure*}[t]
  \centering
   \includegraphics[width=\linewidth]{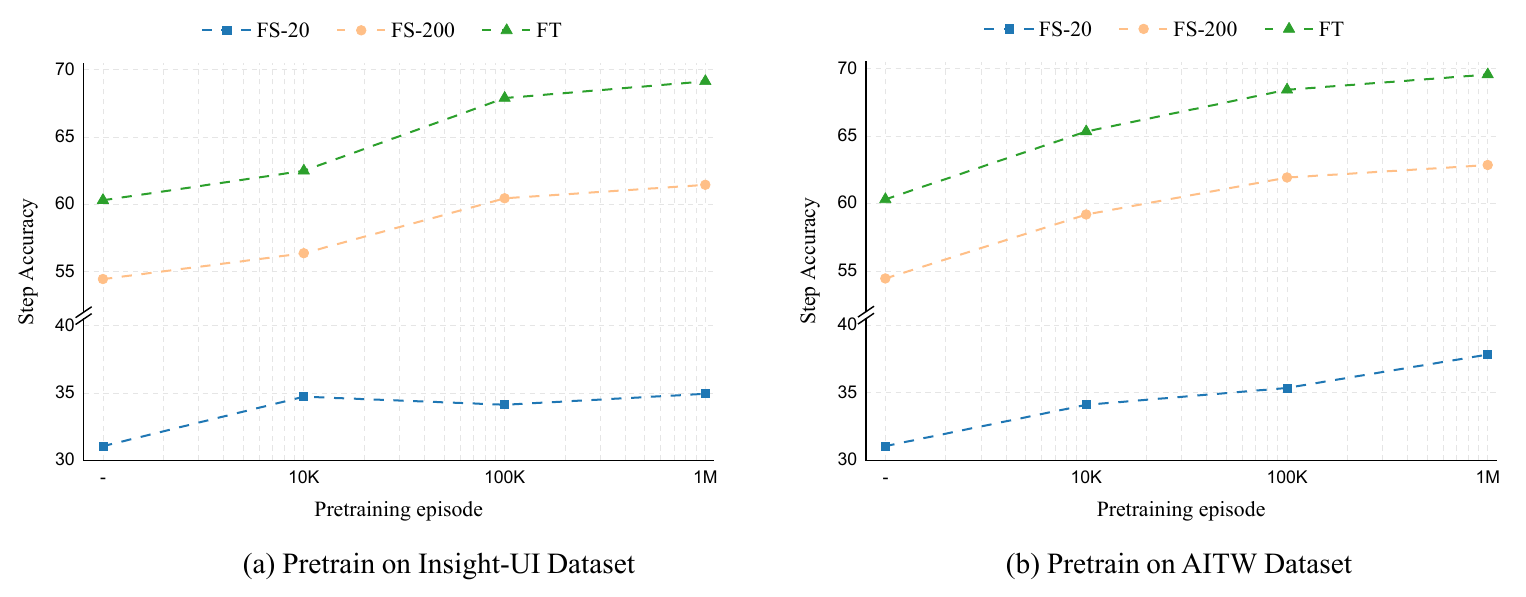}
   \caption{After pretraining on the general dataset (\dataset) and the in-domain dataset (instruction-free AITW), the model is fine-tuned and evaluated on AITZ. ``FS-20" or ``FS-200" indicates that the model is few-shot fine-tuned on 20 or 200 random AITZ training samples, while ``FT" denotes fine-tuning on the entire AITZ training set (1998 samples).}
   \label{fig:AITW_pretrain}
\end{figure*}

\subsection{Ablation Study}

We conduct ablation studies to investigate the effect of different data mixtures on pretraining.
We utilize three types of data: (1) GUI context data, represented by our proposed \dataset; (2) GUI grounding data, consisting of training data from SeeClick, including Web data from SeeClick~\cite{SeeClick} and Mobile data from Widget Captioning~\cite{WidgetCaption}, Rico~\cite{Rico}, and Screen2Words~\cite{Screen2words}; (3) instruction data, including M4-Instruct~\cite{M4-Instruct} and LLaVa’s 158K GPT-generated instruction data~\cite{LLaVA}.
We apply different data mixture strategies to pretrain Qwen2-VL-7B and evaluate its performance on AITZ. The results are shown in  \cref{fig:data_mixture}.

We observe that single-source data leads to suboptimal performance.
Specifically, when only instruction-following data is used, overall GUI performance decreases. This is expected, as there is a domain gap between GUI-specific tasks and general scenarios.
When only GUI grounding data is used, the model shows a modest improvement.
As described in \cite{Qwen2VL}, Qwen2-VL is pretrained to develop basic GUI grounding, therefore, additional pretraining on single-image tasks does not yield significant improvements.
Although \dataset significantly enhances Qwen2-VL's performance, once the pretraining volume exceeds 100K, the improvement rapidly plateaus.

Combining GUI context data and GUI grounding data does not lead to notable improvements to GUI context data only. If a model can navigate to another page, it must also understand the functionality of each element on the page, suggesting that GUI context data partially replace GUI grounding data’s role.
Finally, combining GUI context data with instruction-following data further enhances overall accuracy, especially when the data volume exceeds 100K.
This further supports the notion that GUI context knowledge and instruction-following capability jointly enhance the performance of GUI agents.

\subsection{Further Analysis on Data Domain}

Our proposed \dataset consists of a wide range of Internet websites, representing a general scenario.
AITZ is a cleaned subset of AITW, primarily removing episodes with incorrect actions and providing more detailed and diverse instructions, while all screenshots are still sourced from AITW.
This makes AITW a suitable in-domain pretraining corpus for AITZ.
We convert AITW into our GUI context format by removing all text instructions and randomly cropping observation-action sequences from AITW episodes.
We generate 434K GUI contexts from AITW to make a fair comparison with our \dataset.
After being mixed separately with instruction-following data, the general and in-domain pretraining data are used to pretrain Qwen2-VL-7B, followed by fine-tuning and evaluation on AITZ.
The results are shown in \cref{fig:AITW_pretrain}.

Both general and in-domain pretraining data consistently improve the overall downstream performance of the GUI agent, suggesting that GUI context pretraining effectively alleviates the need for extensive downstream data collection.
For example, when pretrained on 100K GUI context data, the model fine-tuned with 200 downstream samples achieves performance comparable to that of a model directly fine-tuned on 2000 downstream samples.

In-domain pretraining data generally performs better, especially when the downstream data volume is limited.
As the downstream and pretraining data volumes increase, the performance gap narrows.
This suggests a potential solution for improving GUI agents: first pretrain on extensive general data to acquire common GUI knowledge, and then fine-tune on domain-specific data to achieve higher accuracy with an acceptable data volume.

%% file: sec/5_conclusion.tex
\section{Conclusion}
\label{sec:conslusion}

In this paper, we address the two core abilities to develop a GUI agent: understanding GUI scenarios and interpreting user intentions from instructions to predict subsequent actions.
We propose to decouple these two aspects to improve the model's comprehension of GUI environments.
We introduce a fully auto-generated, instruction-free GUI navigation dataset, \dataset, which encompasses diverse GUI context knowledge.
With enhanced GUI context comprehension, \model outperforms previous work across various benchmarks.
In future work, we plan to extend \model by combining general GUI knowledge with specific app perception to achieve a balance between model performance and overall cost.

%% file: sec/X_supplementary.tex
\clearpage
\setcounter{page}{1}
\maketitlesupplementary

\begin{figure}[t]
  \centering
   \includegraphics[width=\linewidth]{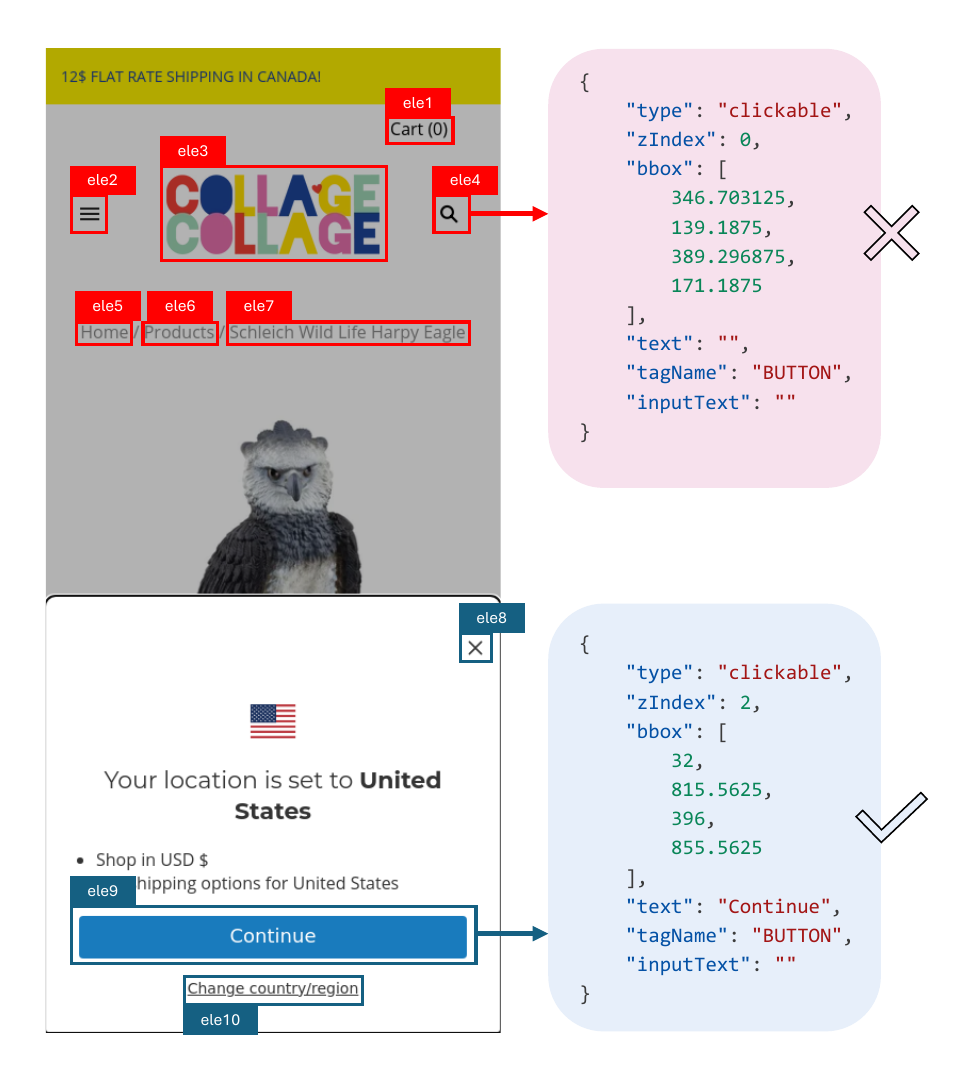}
   \caption{Hierarchical structure is applied in the webpage design, only the upper layer elements will be selected.}
   \label{fig:stack_order}
\end{figure}

\begin{figure}[t]
  \centering
   \includegraphics[width=\linewidth]{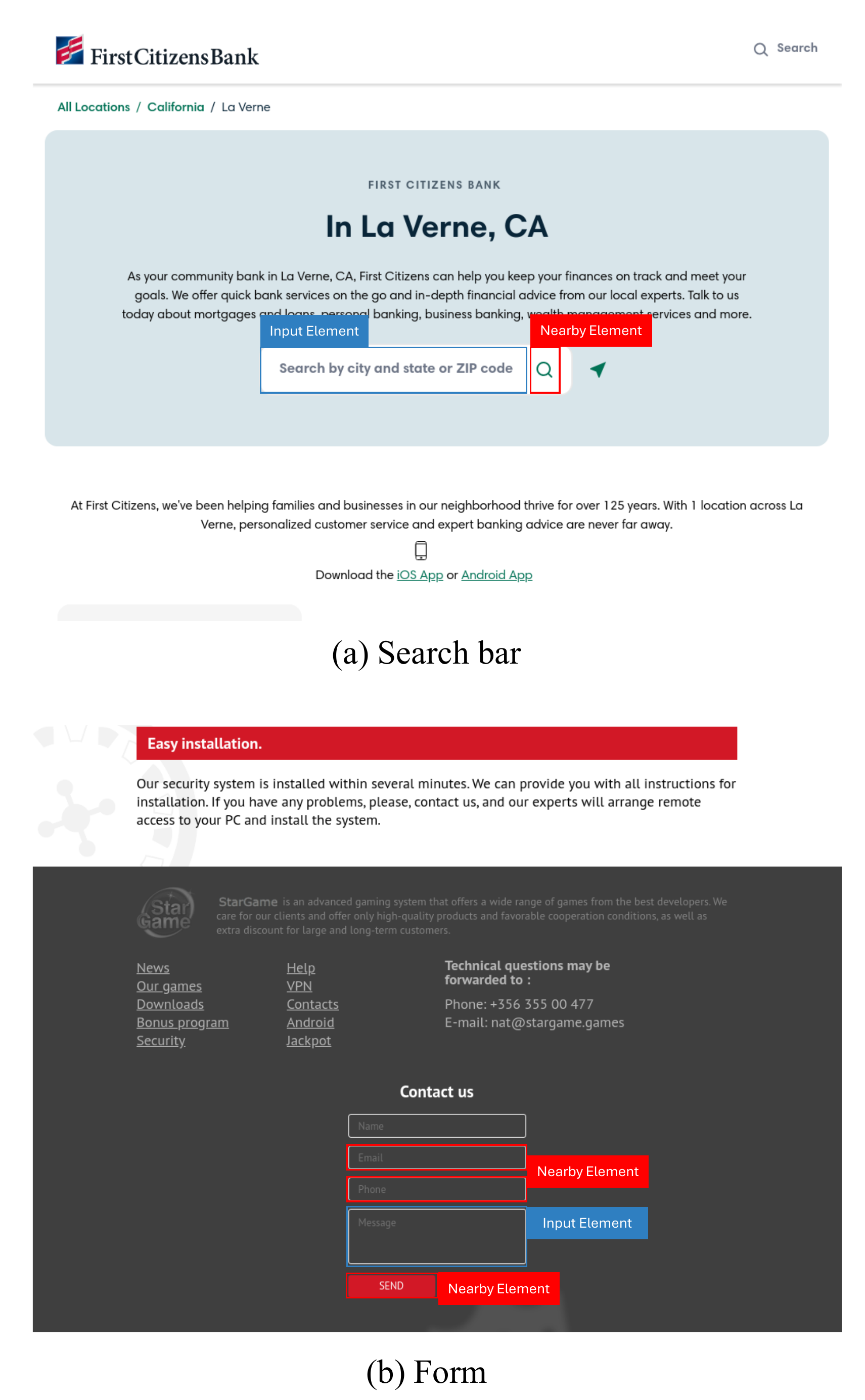}
   \caption{Input elements are predominantly found in search bars and forms. After interacting with these elements, subsequent actions typically involve nearby elements.}
   \label{fig:input_element}
\end{figure}

\section{Action Choosing Rules in Dataset Collection}

As mentioned in \cref{data_construct}, \dataset consists of interleaved observation-action sequences.
However, completely random action selection can result in repetitive or meaningless steps, degrading dataset quality.
Therefore, we apply heuristic rules to refine action selection during dataset construction.

For each page, we first identify all visible elements and retain clickable or inputable ones to create the initial action set.
All refinement operations are performed on this action set.

\begin{figure*}[t]
  \centering
   \includegraphics[width=\linewidth]{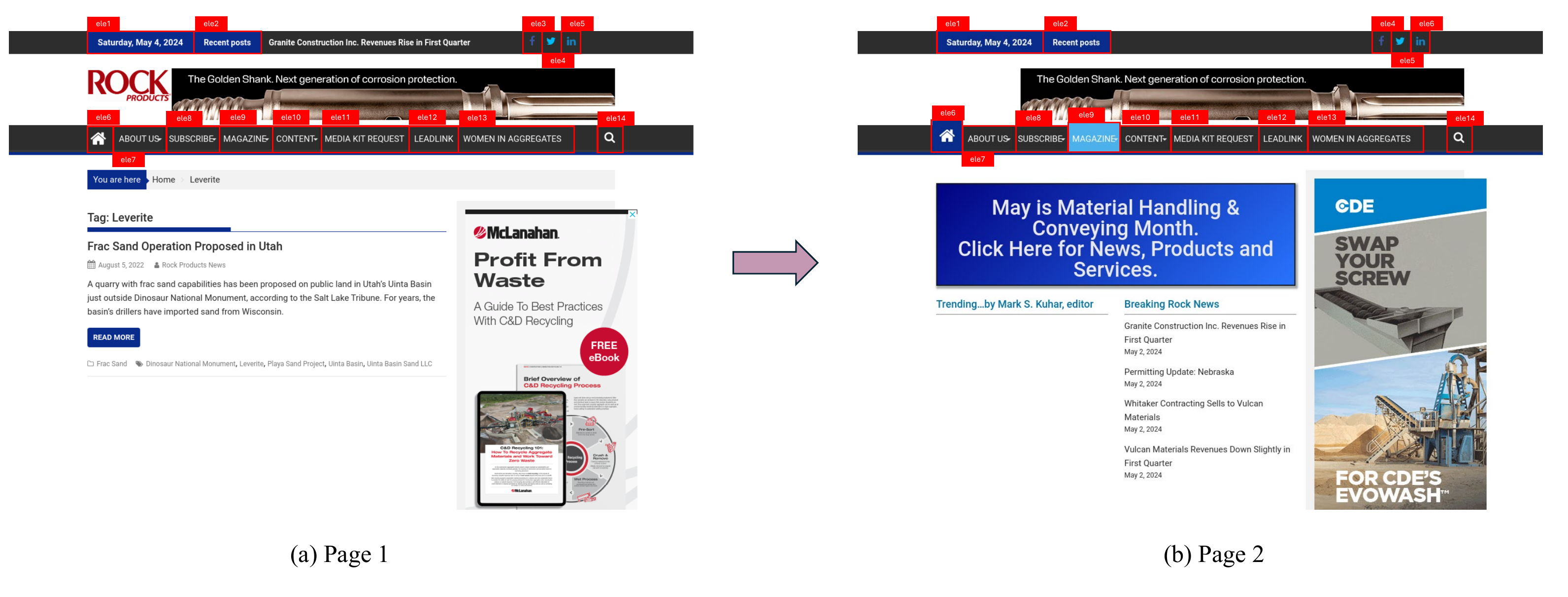}
   \caption{Some elements remain present in subsequent observations after the interaction. These elements lead to the same result regardless of being on Page 1 or Page 2. Therefore, on Page 2, elements shared between the two pages are excluded from the action selection.}
   \label{fig:same_element}
\end{figure*}

\begin{itemize}
    \item The GUI adopts a hierarchical structure, with each page potentially comprising multiple stacked layers.
    As shown in \cref{fig:stack_order}, the page contains two main layers.
    The upper layer is a pop-up prompting the user to provide their location.
    Users typically interact with the pop-up first.
    In HTML, elements in the upper layer have a higher ``z-index" value, so elements with a lower ``z-index" in the action set are removed.
    \item After the interaction, certain elements persist in the new observation. As shown in \cref{fig:same_element}, elements 1 to 14 appear on both Page 1 and Page 2. 
    Interacting with these elements could result in repeated procedures.
    For instance, interacting with element 13 on Page 2 yields the same results as interacting with element 13 on Page 1.
    Therefore, when interacting with Page 2, elements 1 to 14 are excluded from the action set to avoid redundant actions.
    An exception is made only if no new elements are observed on Page 2.
    \item For input actions, we observe that they are primarily performed in search bars and forms, as shown in \cref{fig:input_element}.
    In such cases, most actions following the input action occur nearby. Thus, when an input action is selected, the subsequent action is chosen near the input area.
    Specifically, the input bounding box is expanded using its shorter edge length.
    If an element does not overlap with the expanded bounding box or is separated by more than two elements in the action set, it will be removed from the action set.
    \item Finally, the scroll action is added to the action set, and one action is randomly chosen from the remaining set for execution.
\end{itemize}

\section{Experiment Settings}

\subsection{Training Format}

\begin{figure*}[t]
  \centering
   \includegraphics[width=\linewidth]{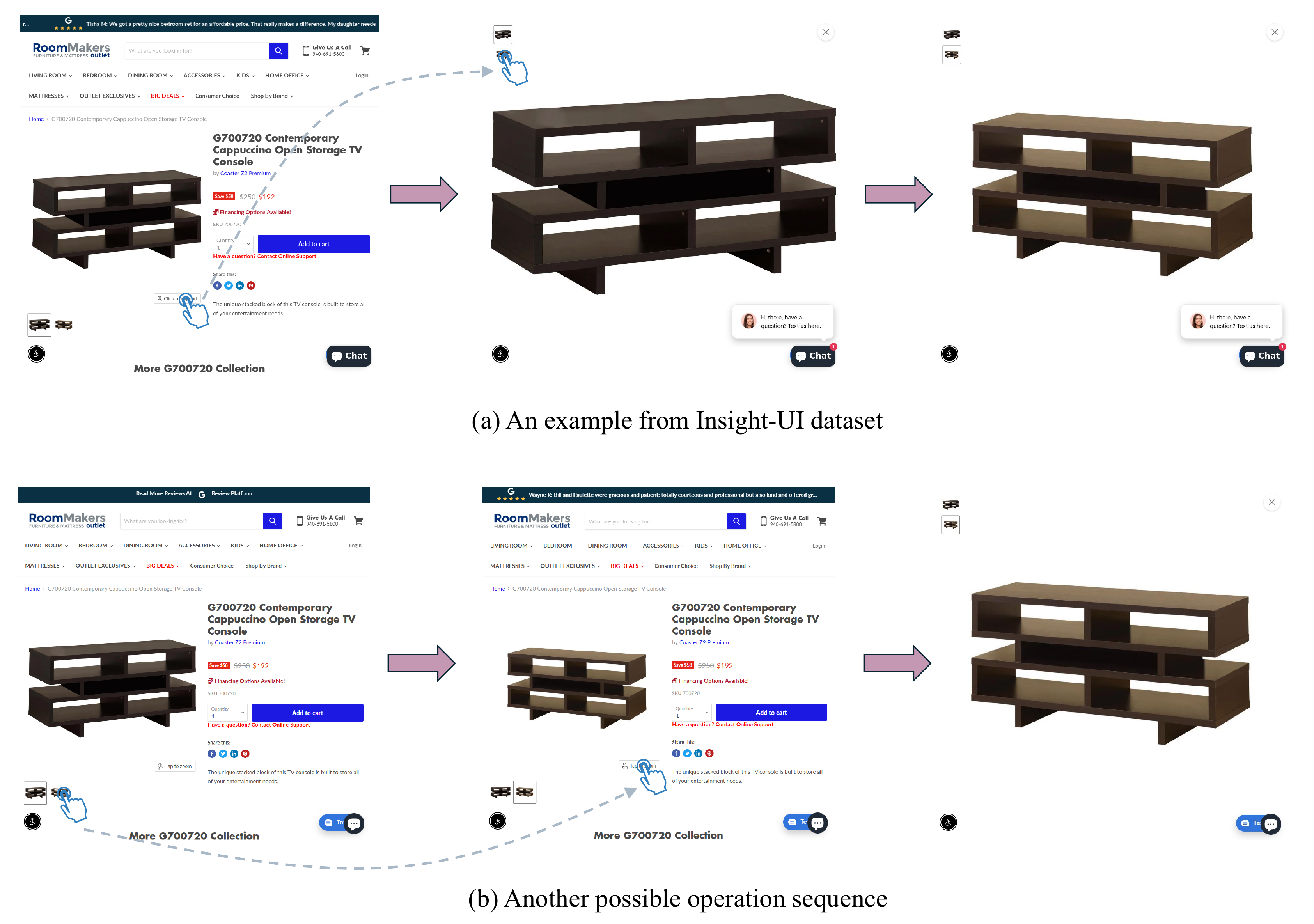}
   \caption{For the same initial and final observations, multiple navigation routes can achieve the target. (a) An example from our proposed Insight-UI dataset. (b) We extract the page URL from the raw data, and a new episode is manually captured to achieve the same result with a different operation sequence.}
   \label{fig:data_split_1}
\end{figure*}

\begin{figure*}[t]
  \centering
   \includegraphics[width=\linewidth]{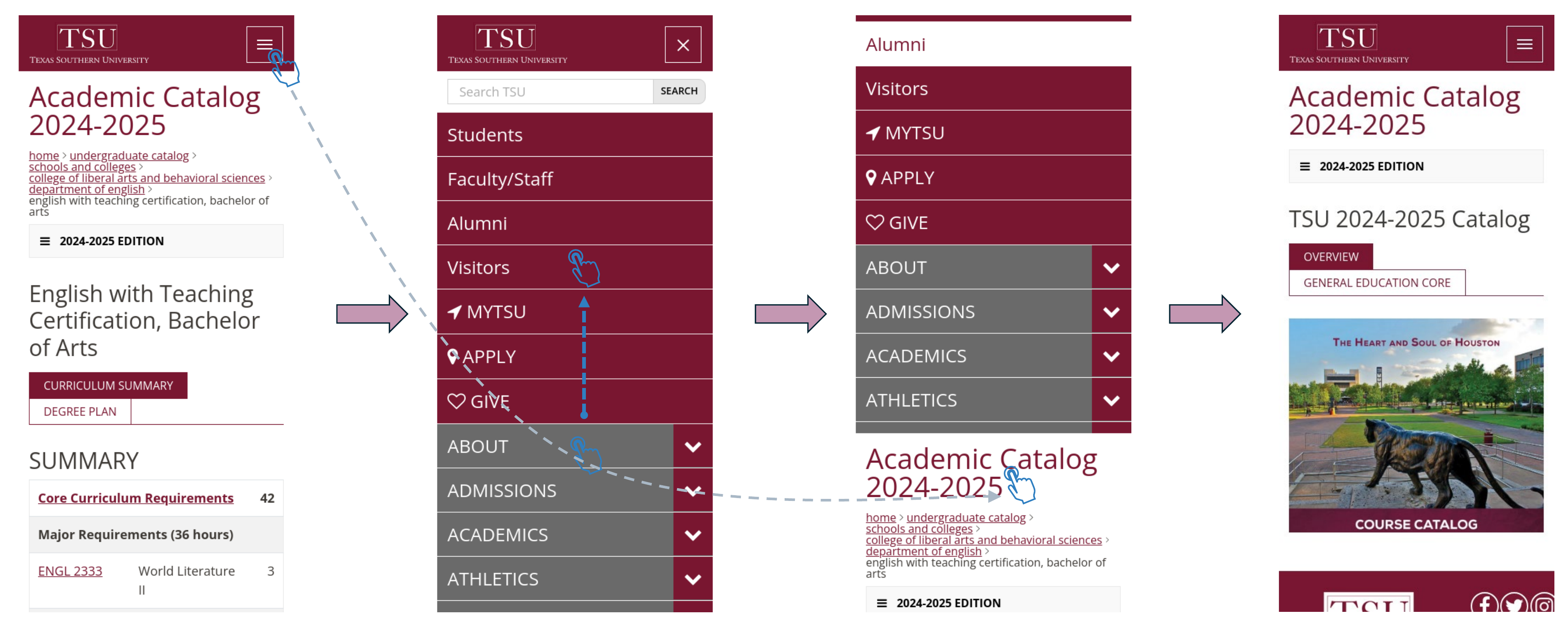}
   \caption{Certain intermediate operations are unnecessary to achieve the target. The initial observation can directly transition to the final observation by clicking the title ``Academic Catalog 2024-2025".}
   \label{fig:data_split_2}
\end{figure*}

\begin{figure*}[t]
  \centering
   \includegraphics[width=\linewidth]{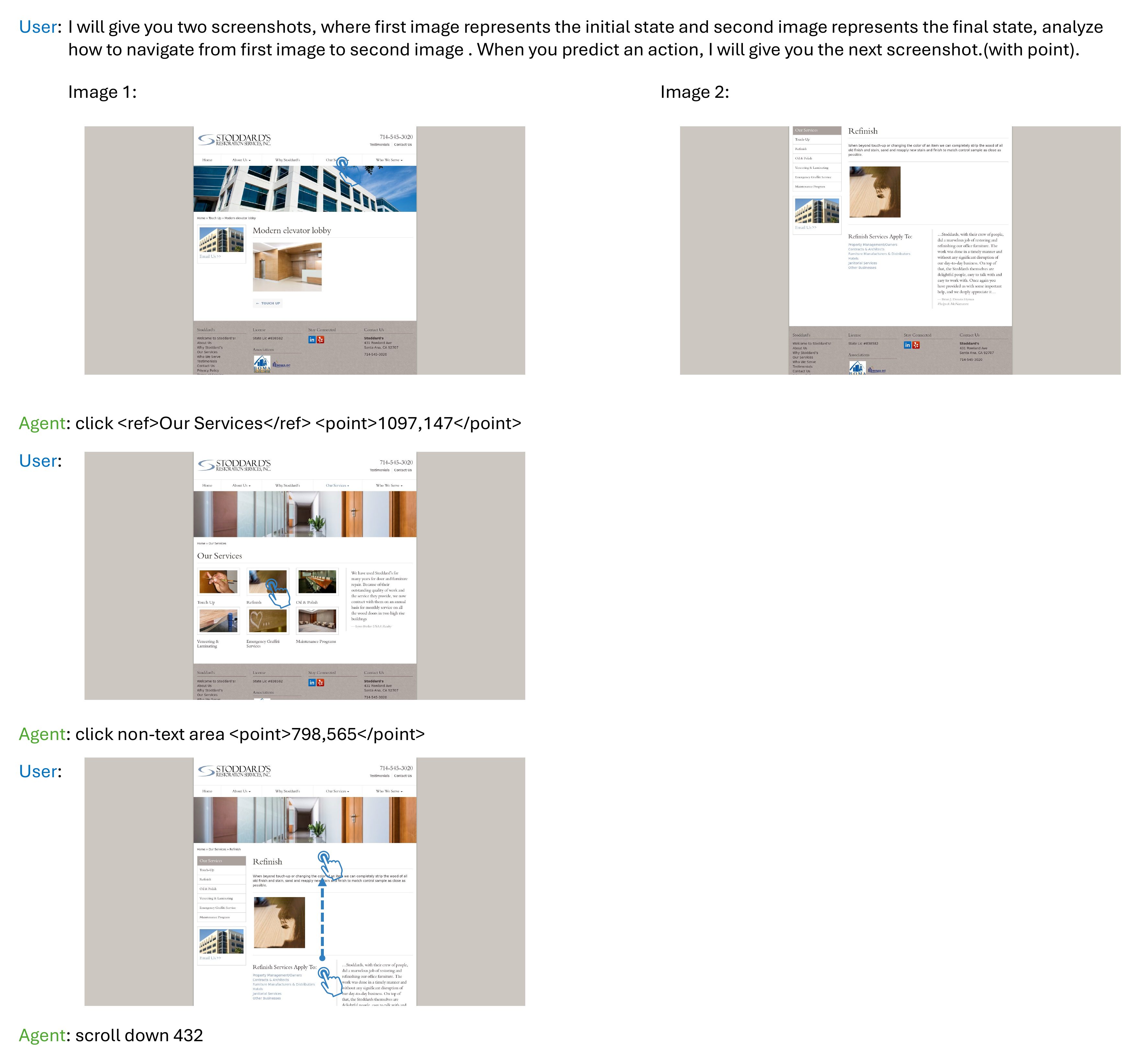}
   \caption{The first pretraining format. In this format, only the initial and final observations are provided. Each time the model predicts an action, a new observation is offered, simulating real interaction logic in downstream GUI agent tasks.}
   \label{fig:format1}
\end{figure*}

\begin{figure*}[t]
  \centering
   \includegraphics[width=\linewidth]{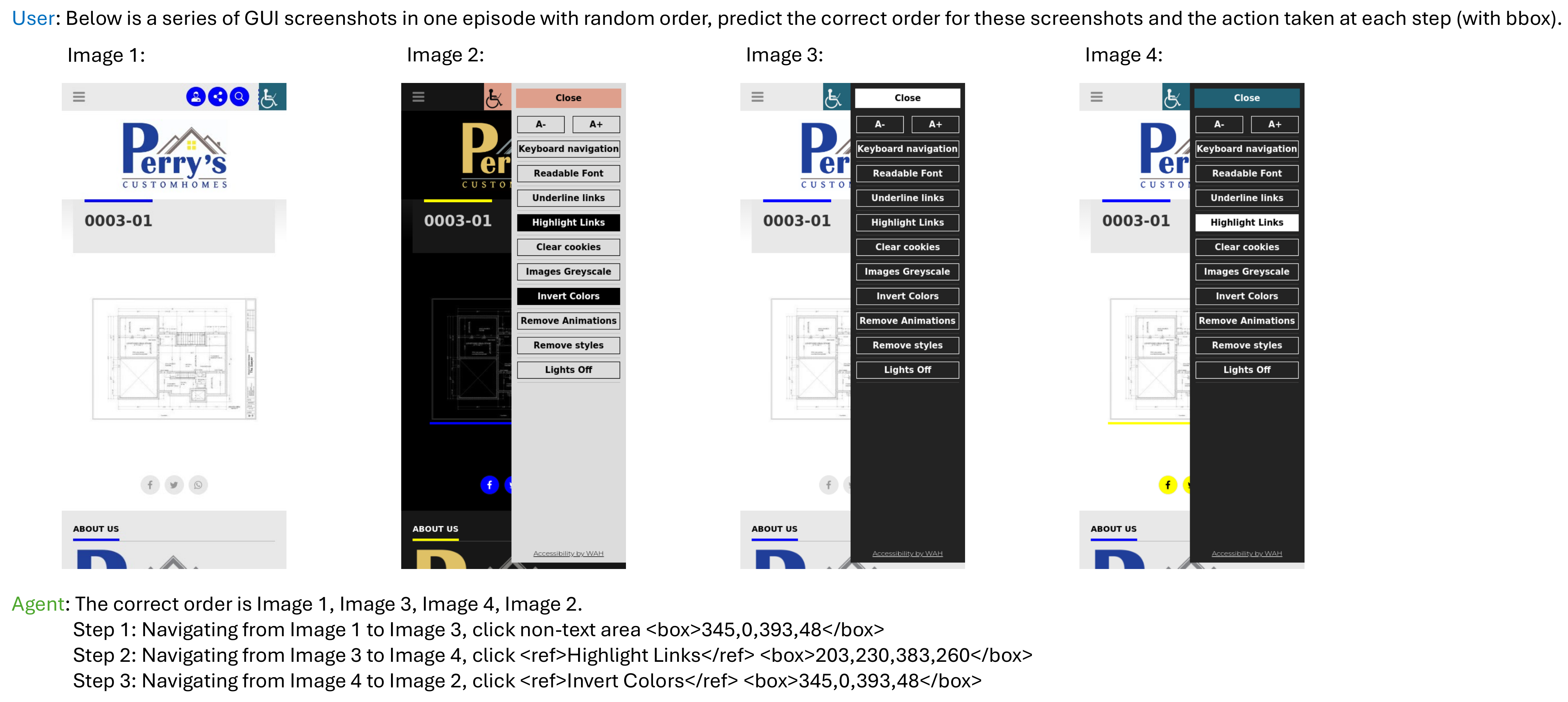}
   \caption{The second pretraining format. In this format, all observations are initially provided, and the model is required to determine how to transition between observations using a single action directly.}
   \label{fig:format2}
\end{figure*}

Despite applying several data processing procedures to ensure quality, directly requiring the model to navigate from the first to the last observation remains challenging.

First, there may be multiple routes for the same examples.
For instance, \cref{fig:data_split_1}(a) presents an example from the Insight-UI dataset, where the process first expands the item image and then switches perspectives.
However, alternative operations can achieve the same result, as illustrated in \cref{fig:data_split_1}(b).

Additionally, some operations may be redundant.
As illustrated in \cref{fig:data_split_2}, the final observation can be reached directly from the initial observation by clicking the title ``Academic Catalog 2024-2025".
Intermediate operations have no impact on the final results.

These issues are challenging to eliminate during data collection, and using LVLMs for cleaning is both costly and imprecise.
Unlike downstream GUI agent tasks, GUI context pretraining focuses on teaching the model navigation logic in GUI scenarios rather than adapting to specific apps or websites.
This enables splitting the original long sequences into several shorter ones.
To address this issue, we divide the pretraining tasks into two formats.

Our proposed GUI context pretraining and GUI Agent tasks are modeled as multi-turn, multi-modal conversations.
\cref{fig:format1} illustrates the original training format for GUI context, which closely resembles the downstream GUI agent tasks, except that text instructions are replaced with final observations.
Using this format, we first train the model on 2,000 randomly selected examples from \dataset, and then evaluate it on examples requiring more than one action from the entire dataset.
During evaluation, the model input is ground truth, applying a teacher-forcing strategy similar to training.
The average loss of the model's predicted tokens is calculated for each episode.
Finally, half of the examples with higher evaluation losses are selected for the second pretraining format, as depicted in \cref{fig:format2}.

\subsection{Training Configuration}

Our model is initialized from Qwen2-VL-7B and continually pretrained using the data described in \cref{model_optimization}.
We follow the official code in Qwen2-VL\footnote{https://github.com/QwenLM/Qwen2-VL} and LLaMA-Factory\footnote{https://github.com/hiyouga/LLaMA-Factory}.
Preliminary experiments show that finetuning the vision encoder in Qwen2-VL does not yield consistent performance improvements. Therefore, the vision encoder is frozen in subsequent experiments.
LoRA is applied to fine-tune the model.
AdamW is adopted as the optimizer with a learning rate of 1e-4.
The model undergoes linear warm-up for the first 10\% of steps, followed by cosine decay.
The global batch size is configured as 64.
All experiments are conducted on 8 NVIDIA A800 GPUs (80 GB).

Qwen2-VL can process images of any resolution. Therefore, we only constrain the maximum resolution for input images. Images with resolutions below the limit remain unmodified.
During pretraining, DeepSpeed ZeRO2 is used for the first 80\% of steps, with the vision resolution limited to 840.
For the remaining 20\% of steps, DeepSpeed ZeRO3 is employed, with the vision resolution limited to 1540.
According to SeeClick \cite{SeeClick}, during pretraining, predicting points was slightly better than bounding boxes. So in the \dataset, 70\% of the examples are randomly chosen to predict center point coordinates, while the remaining 30\% are designated for bounding box prediction

\section{Visualization}

\begin{figure*}[t]
  \centering
   \includegraphics[width=0.6\linewidth]{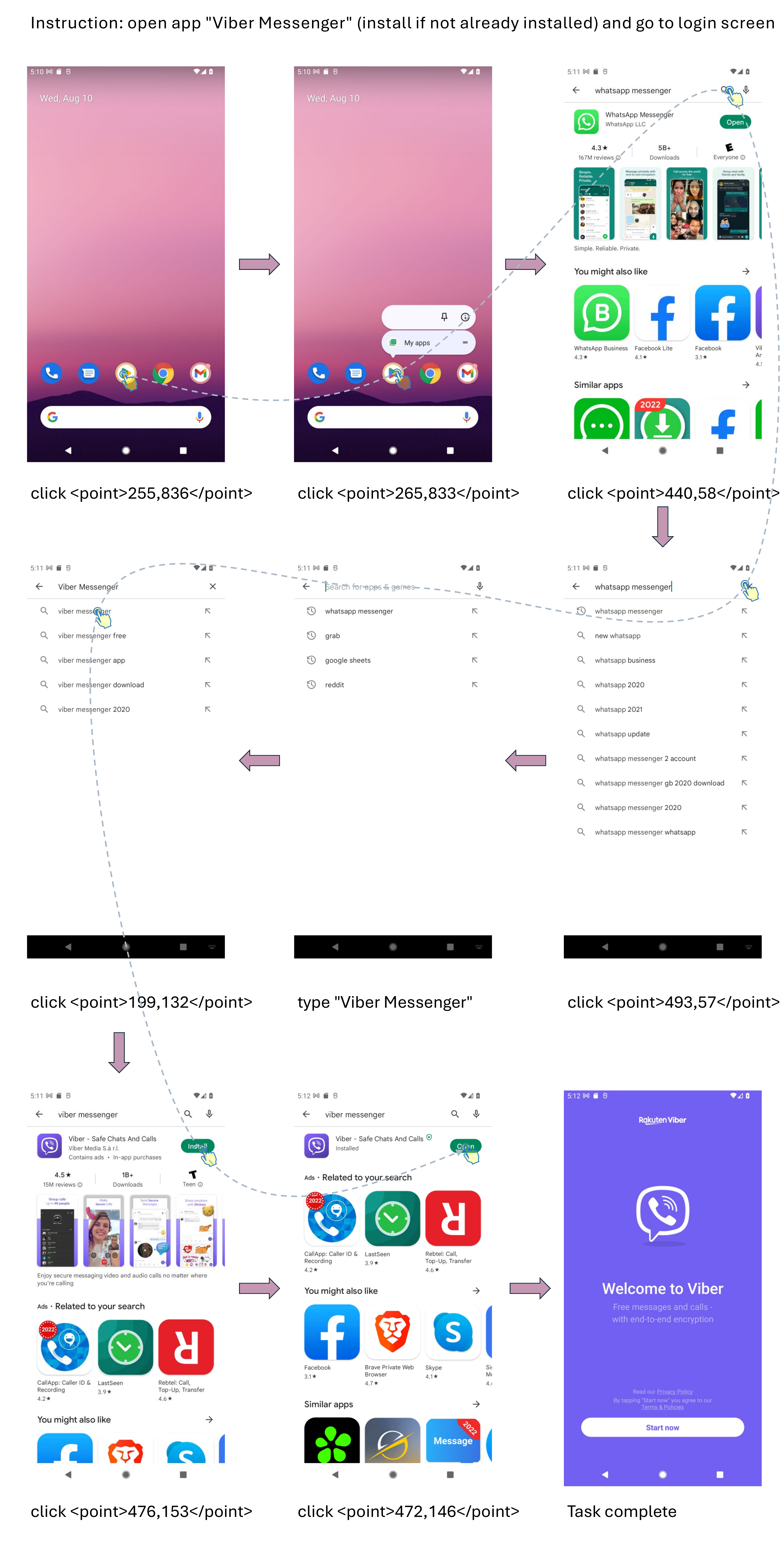}
   \caption{Example episodes of \model on mobile scenarios (AITZ).}
   \label{fig:visualize_android}
\end{figure*}

\begin{figure*}[t]
  \centering
   \includegraphics[width=0.8\linewidth]{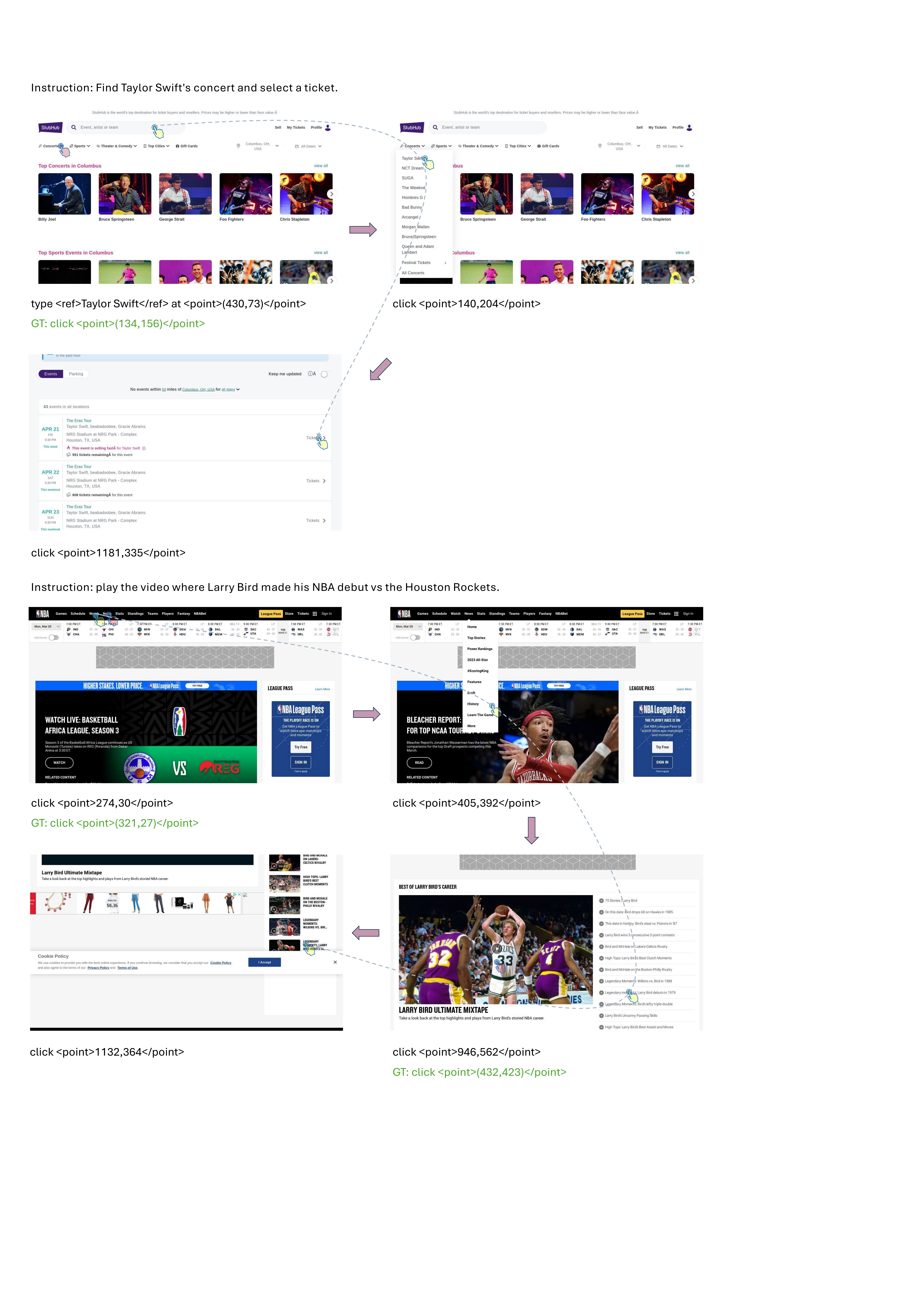}
   \caption{Example episodes of \model in desktop scenarios (Mind2Web). If the model produces an incorrect prediction, the ground truth is displayed below and highlighted within the image in \textcolor{red}{red}. Due to the complexity of real GUI scenarios, offline metrics are not always accurate. For instance, in the first example, \model chooses to search for ``Taylor Swift" in the search bar, which can also yield correct results.}
   \label{fig:visualize_desktop}
\end{figure*}

We visualize examples from both mobile and desktop scenarios to provide a clearer understanding of \model.
\cref{fig:visualize_android} illustrates \model predictions on mobile devices.
\cref{fig:visualize_desktop} displays \model predictions on desktop devices.
\model demonstrates strong generalization across diverse scenarios.